\documentclass[journal]{IEEEtran}

\ifCLASSINFOpdf
\else
\fi

\usepackage{graphicx}
\usepackage{multirow}
\usepackage{subfigure}
\usepackage{amsfonts}
\usepackage[colorlinks,linkcolor=red]{hyperref}
\usepackage{color}
\usepackage{amsmath,amssymb} 

\begin{document}
\title{Shadow Removal by a Lightness-Guided Network with Training on Unpaired Data}

\author{Zhihao~Liu,
        Hui~Yin,
        Yang~Mi,
        Mengyang~Pu,
        and~Song~Wang,~\IEEEmembership{Senior Member,~IEEE}
\thanks{Zhihao~Liu, Hui~Yin, and Mengyang~Pu are with the School of Computer and Information Technology, Beijing Jiaotong University, Beijing 100044 China (e-mail: {16120394,hyin,mengyangpu}@bjtu.edu.cn).}
\thanks{Yang~Mi is with the Department of Computer Science and Engineering, University of South Carolina, Columbia, SC 29201 USA (e-mail: miy@email.sc.edu).}
\thanks{Song~Wang is with the Department of Computer Science and Engineering, University of South Carolina, Columbia, SC 29201 USA, and also with the College of Intelligence and Computing, Tianjin University, Tianjin 300072, China (e-mail: songwang@cec.sc.edu).}}


\markboth{}%
{Liu \MakeLowercase{\textit{et al.}}: Shadow Removal by a Lightness-Guided Network with Training on Unpaired Data}

\maketitle

\begin{abstract}
Shadow removal can significantly improve the image visual quality and has many applications in computer vision. Deep learning methods based on CNNs have become the most effective approach for shadow removal by training on either paired data, where both the shadow and underlying shadow-free versions of an image are known, or unpaired data, where shadow and shadow-free training images are totally different with no correspondence. In practice, CNN training on unpaired data is more preferred given the easiness of training data collection. In this paper, we present a new Lightness-Guided Shadow Removal Network (LG-ShadowNet) for shadow removal by training on unpaired data. In this method, we first train a CNN module to compensate for the lightness and then train a second CNN module with the guidance of lightness information from the first CNN module for final shadow removal. We also introduce a loss function to further utilise the colour prior of existing data. Extensive experiments on widely used ISTD, adjusted ISTD and USR datasets demonstrate that the proposed method outperforms the state-of-the-art methods with training on unpaired data. \footnote{All codes and results are available at \url{https://github.com/hhqweasd/LG-ShadowNet}.}
\end{abstract}


\IEEEpeerreviewmaketitle

\section{Introduction}
\IEEEPARstart{S}{hadow} is a common natural phenomenon and it occurs in regions where the light is blocked. 
Shadow regions are usually darker with insufficient illumination and bring further complexities and difficulties to many computer vision tasks such as semantic segmentation, object detection, and object tracking~\cite{cucchiara2003detecting,jung2009efficient,nadimi2004physical,sanin2010improved}. 
Although many \emph{shadow removal} methods have been developed to recover the illumination in shadow regions, their performance is compromised given the difficulty to distinguish shadows and some darker non-shadow regions. 

\begin{figure}[htbp]
    \centering
    \includegraphics[width=1\linewidth]{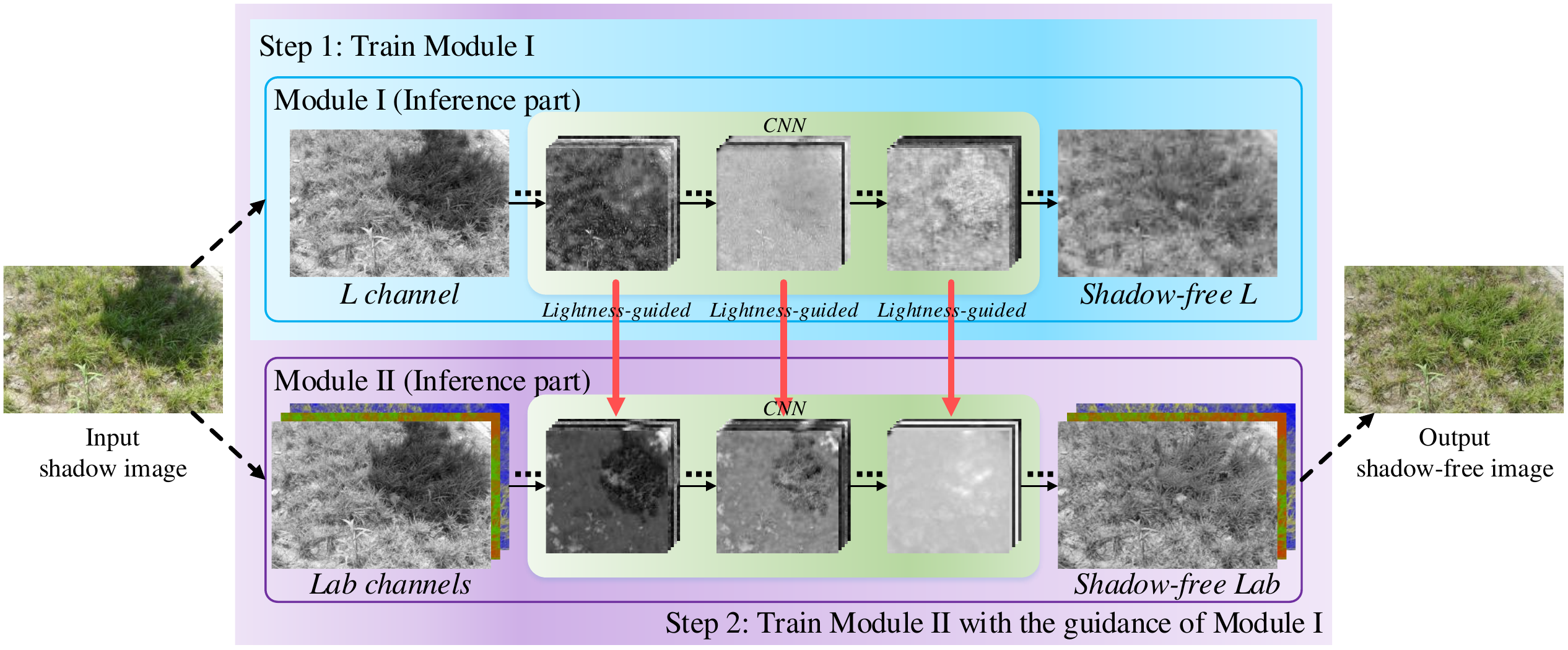}
    \caption{An illustration of the proposed idea of training the shadow removal model with the guidance of lightness information. 
    The first CNN module (Module~\uppercase\expandafter{\romannumeral1}) is trained in the first step for lightness features, which are then connected to the second CNN module (Module~\uppercase\expandafter{\romannumeral2}) in the second step for guiding the learning of shadow removal (red arrows), by further considering all colour information.
    Only the inference part of each module is shown for better looking.}
    \label{fig:1}
\end{figure}

Compared with traditional methods~\cite{finlayson2005removal,guo2012paired,khan2015automatic,zhang2015shadow}, deep learning methods based on convolutional neural networks (CNNs) have been shown to be much more effective for shadow removal by training on annotated data. One popular approach is to use \emph{paired data}, \textit{i.e.}, both the shadow and shadow-free versions of an image, to train CNNs~\cite{ding2019argan,hu2019direction,qu2017deshadownet,wang2018stacked}.
However, it is difficult and time-consuming to collect such image pairs -- it usually requires a highly-controlled setting of the lighting sources, occluding objects, and cameras, as well as a strictly static scene. Data collected in such a controlled setting lacks diversity and the trained CNNs may not handle well general images of different scenes. 

In~\cite{hu2019mask}, Mask-ShadowGAN is proposed for shadow removal by training CNNs on \emph{unpaired data}, where the shadow images and shadow-free images used for training have no correspondence, \textit{i.e.}, they may be taken at different scenes.
Its basic idea is to transform shadow removal to image-to-image translation, based on adversarial learning and cycle-consistency. Clearly, we can collect large-scale, diverse unpaired data easily, which can train CNNs with better generalisation capability for shadow removal.
However, the performance of this method is still inferior to the CNNs trained on paired data, when testing on several well-known benchmark datasets.
The main reason is that it is especially difficult for the model to train on unpaired data from scratch with only image-level annotations, comparing with models trained on paired data with pixel-level annotations.

In this paper we aim to improve the performance of Mask-ShadowGAN~\cite{hu2019mask} by developing a new lightness-guided network with training on unpaired data.
The basic idea of our method is to simplify the unpaired-data learning into two steps: first learn a part of simple and obvious knowledge of shadow, and then use it to guide the learning of whole knowledge of shadow.
In natural scenes, natural light is the main source of illumination, which plays the key role in shadow removal~\cite{finlayson2002removing,finlayson2005removal} and shadow modellings~\cite{Le2019Shadow,shor2008shadow}. In most cases, shadow regions show similar chromaticity as but lower lightness than non-shadow regions\cite{cha99statistical,cucc2001improve}.
Therefore, lightness is a very important cue of shadow regions and following the above two-step idea, we propose to learn the shadow knowledge first and then use it to guide the full learning for shadow removal.
We explore the lightness at feature levels instead of input level to better represent the shadow knowledge by distinguishing the shadow regions and the dark albedo material regions that also show lower lightness and can be easily confused with shadows~\cite{le2018adnet}.

More specifically, by representing the input image in the $Lab$ colour space~\cite{tkalcic2003color}, where $L$ channel reflects the image lightness, we first train a CNN module (Module~\uppercase\expandafter{\romannumeral1}) to compensate for the lightness in the $L$ channel. 
As illustrated in Fig.~\ref{fig:1}, we propose to use the learned CNN features of lightness to help train a second CNN module (Module~\uppercase\expandafter{\romannumeral2}) for shadow removal by considering all $Lab$ channels.
The first CNN module is connected to the second CNN module through multiplicative connections~\cite{feichtenhofer2017spatiotemporal} to form a Lightness-Guided Shadow Removal Network (LG-ShadowNet).
The multiplicative connection can combine the features of one stream with the features of another stream, \textit{i.e.}, combine the lightness features from Module~\uppercase\expandafter{\romannumeral1} with the features of Module~\uppercase\expandafter{\romannumeral2}, leading to a two-stream-like lightness-guided architecture.

Furthermore, we introduce a new loss function to further utilise the colour prior of existing data. This loss is a variant of the colour loss used for image enhancement~\cite{wang2019under}, which encourages the learning of colour consistency between the generated data and the input data.
Considering the performance and computational efficiency, we keep the number of parameters of LG-ShadowNet roughly the same as Mask-ShadowGAN~\cite{hu2019mask} by following the strategies used for SqueezeNet~\cite{iandola2016squeezenet}.
In the experiments, we also discuss the use of value channel in HSV colour space for lightness feature learning given its similarity to the lightness channel in $Lab$ colour space.

The main contributions of this work are:
\begin{itemize}
    \item A new lightness-guided method is proposed for shadow removal by training on unpaired data. It fully explores the important lightness information by first training a CNN module only for lightness before considering other colour information.
    \item An LG-ShadowNet is proposed to integrate the lightness and colour information for shadow removal through multiplicative connections. We also explore various alternatives for these connections and introduce a new loss function based on colour priors to further improve the shadow removal performance.
    \item Extensive experiments are conducted on widely used ISTD~\cite{wang2018stacked}, adjusted ISTD~\cite{Le2019Shadow} and USR~\cite{hu2019mask} datasets to validate the proposed method as well as justifying its main components.
    Experimental results demonstrate that the proposed method outperforms the state-of-the-art methods with training on unpaired data.
\end{itemize}

\section{Related Work}

In this section, we briefly review the related work on shadow removal and two-stream CNN networks. 

\subsection{Shadow removal}
Traditional shadow removal methods use gradient~\cite{gryka2015learning}, illumination~\cite{zhang2015shadow,shor2008shadow,xiao2013fast}, and region~\cite{guo2012paired,vicente2017leave} information to remove shadows. 
In recent years, deep learning methods based on CNNs have been developed for shadow removal with significantly better performance than the traditional methods. 
Most of them rely on paired data for supervised training. 
In~\cite{qu2017deshadownet}, a multi-context architecture is explored to embed information from three different perspectives for shadow removal, including global localisation, appearance, and semantics. 
In~\cite{wang2018stacked}, a stacked conditional generative adversarial network is developed for joint shadow detection and shadow removal. 
In~\cite{hu2019direction}, direction information is further considered to improve shadow detection and removal.
In~\cite{ding2019argan}, an attentive recurrent generative adversarial network is proposed to detect and remove shadows by dividing the task into multiple progressive steps. 
In~\cite{Le2019Shadow}, a shadow image decomposition model is proposed for shadow removal, which uses two deep networks to predict unknown shadow parameters and then obtain the shadow-free image according to their decomposition model.
To remove the reliance on paired data, a Mask-ShadowGAN framework~\cite{hu2019mask} is proposed based on the cycle-consistent adversarial network of CycleGAN~\cite{zhu2017unpaired}.
By introducing the generated shadow masks into CycleGAN, Mask-ShadowGAN uses unpaired data to learn the underlying mapping between the shadow and shadow-free domains. 
However, these CNN-based methods process all information (or all channels) of the input images together, and none of them takes out the lightness information from the input images for a separate training.
In this paper, we train a CNN module exclusively for lightness before considering other colour information and the proposed
LG-ShadowNet trained on unpaired data can achieve comparable performance as the state-of-the-art CNN methods trained on paired data.

\begin{figure*}[htbp]
    \centering
    \includegraphics[width=1\linewidth]{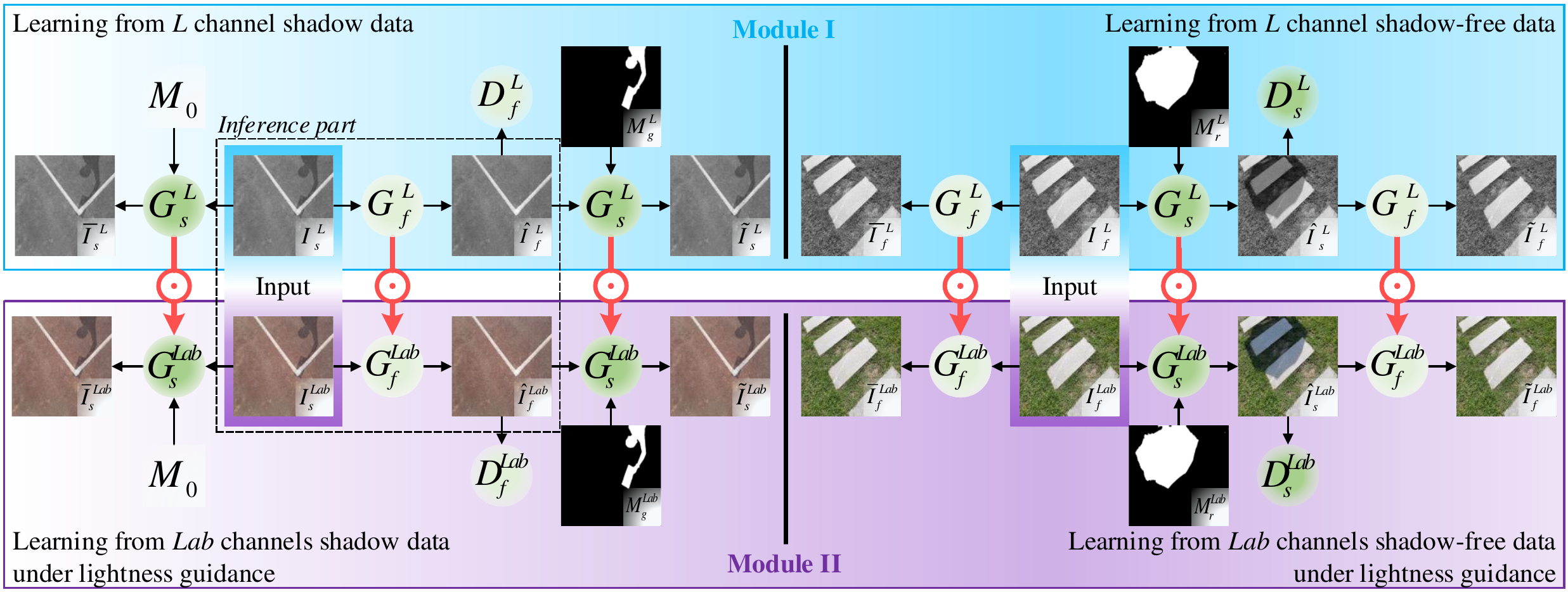}
    \caption{An overview of the proposed LG-ShadowNet. From top to bottom show its two CNN modules, and left and right illustrate the learning from the shadow data and the shadow-free data, respectively. The generators of two modules are connected through multiplicative connections (red arrows with $\odot$).
    The input and output data are shown as 8-bit images rather than separate channels for easy understanding.}
    \label{fig:model}
\end{figure*}
\subsection{Two-stream architecture}
Our lightness-guided architecture are derived from two-stream architectures which have been successfully used for solving many computer vision and pattern recognition tasks~\cite{feichtenhofer2017spatiotemporal,feichtenhofer2016convolutional,gundogdu2019garnet,gupta2007two,peng2016multi,shi2019two,simonyan2014two,zhou2017two,zhou2018learning}.
In~\cite{gupta2007two}, a two-stream processing technique is proposed to fuse the acoustic features and semantics of the conversation for emotion recognition. 
In~\cite{feichtenhofer2016convolutional,simonyan2014two}, two-stream CNN architectures are developed to combine the spatial and temporal information for video-based action recognition. 
In~\cite{shi2019two}, a two-stream framework is proposed to combine the first-order and the second-order information of skeleton data for action recognition. 
In~\cite{gundogdu2019garnet}, a two-stream network is developed to extract garment and 3D body features, which are fused for 3D cloth draping.
In~\cite{feichtenhofer2017spatiotemporal}, motion gating is employed to the residual connections in a two-stream CNN, which can benefit action recognition.
While the lightness-guided architecture of the proposed LG-ShadowNet is structurally similar to the one used in~\cite{feichtenhofer2017spatiotemporal}, they solve completely different problems: shadow removal in this paper and action recognition in~\cite{feichtenhofer2017spatiotemporal}.
Furthermore, in our LG-ShadowNet, the two modules work like a teacher (Module~\uppercase\expandafter{\romannumeral1}) and a student (Module~\uppercase\expandafter{\romannumeral2}) for lightness guidance and shadow removal respectively, while the two streams in~\cite{feichtenhofer2017spatiotemporal} work simply like teammates with the same goal.

\section{Methodology}
In this section, we first elaborate on the proposed LG-ShadowNet and loss function, and then sketch network details of LG-ShadowNet.
After that, we describe the details of multiplicative connections.
Finally, we analyse the convergence of LG-ShadowNet.

\subsection{Proposed network}
The pipeline of the proposed LG-ShadowNet that consists of two modules are shown in Fig.~\ref{fig:model}.
We first train Module~\uppercase\expandafter{\romannumeral1} as shown in the top of Fig.~\ref{fig:model} for lightness compensation, which learns a mapping between the shadow domain and the shadow-free domain on the $L$ channel of $Lab$ images.

In Module~\uppercase\expandafter{\romannumeral1}, generator $G^L_{f}$ maps shadow data $I^L_s$ to shadow-free data $\hat{I}^L_f$, which is further mapped to shadow data $\widetilde{I}^L_s$ by generator $G^L_{s}$, as illustrated in the top-left of Fig.~\ref{fig:model}.
Generator $G^L_{s}$ maps shadow-free data $I^L_f$ to shadow data $\hat{I}^L_s$, which is further mapped to shadow-free data $\widetilde{I}^L_f$ by generator $G^L_{f}$, as illustrated in the top-right of Fig.~\ref{fig:model}.
In these processes, masks $M^L_g$ and $M^L_r$ are used to guide the shadow generation and are computed by following~\cite{hu2019mask}.
Note that $M^L_r$ is a randomly selected mask from the previous computed $M^L_g$. 
Discriminators $D^L_s$ and $D^L_f$ are introduced to distinguish $\hat{I}^L_s$ and $\hat{I}^L_f$ from $I^L_s$ and $I^L_f$, respectively.
$G^L_{f}$ also maps shadow-free data $I^L_f$ to shadow-free data $\overline{I}^L_f$, and $G^L_{s}$ also maps shadow data $I^L_s$ to shadow $\overline{I}^L_s$ with the guide of all-zero-element shadow-free mask $M_0$.

The training of Module~\uppercase\expandafter{\romannumeral1} is the same as the training of Mask-ShadowGAN~\cite{hu2019mask}.
When the training is finished, we fix its parameters and move on to the learning of Module~\uppercase\expandafter{\romannumeral2}, as shown in the bottom of Fig.~\ref{fig:model}.
The parameters of generators of Module~\uppercase\expandafter{\romannumeral2} is initialised with the parameters of Module~\uppercase\expandafter{\romannumeral1} except for the input and output layers.
Module~\uppercase\expandafter{\romannumeral2} is connected with the Module~\uppercase\expandafter{\romannumeral1} by multiplicative connections, resulting in the overall architecture of LG-ShadowNet. 
Module~\uppercase\expandafter{\romannumeral2} takes all $Lab$ channels as input and finalise the shadow removal.

Actually, after connecting $G^L_{f}$ to $G^{\mathit{Lab}}_{f}$, we get a new combined generator $G_f$.
Similarly, the connection of $G^L_{s}$ and $G^{\mathit{Lab}}_{s}$ also 
leads to a new combined generator $G_s$.
When learning from shadow data as shown in the left of Fig.~\ref{fig:model}, $G_f$ converts shadow data $I_s=(I^{L}_{s}, I^{\mathit{Lab}}_{s})$ to the shadow-free data $\hat{I}_f =(\hat{I}^{L}_f, \hat{I}^{\mathit{Lab}}_f)$ and $I_s$ is from a shadow-data distribution $p(I_s)$.
$G_s$ converts the shadow-free data $\hat{I}_f$ to a generated $Lab$ shadow image $\widetilde{I}^{\mathit{Lab}}_{s}$ with the guide of the shadow mask pair $M_g=(M^L_g, M^{\mathit{Lab}}_g)$ that are computed from Modules \uppercase\expandafter{\romannumeral1} and \uppercase\expandafter{\romannumeral2}, respectively. 
This whole learning process can be summarised as
\begin{equation}
    \widetilde{I}^{\mathit{Lab}}_{s}=G_{s}(G_{f}(I_s),M_g).
\end{equation}
The discriminator $D^{\mathit{Lab}}_f$ is introduced to distinguish $\hat{I}^{\mathit{Lab}}_f$ from $I^{\mathit{Lab}}_{f}$.
$G_{s}$ maps $I_s$ to $\overline{I}^{\mathit{Lab}}_s$ with the guide of shadow-free mask $M_0$.

When learning from shadow-free data as shown in the right of Fig.~\ref{fig:model}, the input $I_f=(I^{L}_{f}, I^{\mathit{Lab}}_{f})$ is from a shadow-free data distribution $p(I_f)$.
$G_f$ converts $I_f$ to a shadow data $\hat{I}_s$ with the guide of the mask pair $M_r=(M^L_r, M^{\mathit{Lab}}_r)$, which are randomly selected from the previous obtained $M_g$.
$G_f$ converts $\hat{I}_s$ to a generated $Lab$ shadow-free image $\widetilde{I}^{Lab}_{f}$:
\begin{equation}
    \widetilde{I}^{Lab}_{f}=G_{f}(G_{s}(I_f,M_r)).
\end{equation}
Discriminator $D^{\mathit{Lab}}_s$ is utilised to distinguish $\hat{I}^{\mathit{Lab}}_s$ from $I^{Lab}_{s}$ and $G_{f}$ maps $I_f$ to $\overline{I}_s$.

In short, the training of LG-ShadowNet can be briefly described as two steps: first train Module \uppercase\expandafter{\romannumeral1} and then train Modules \uppercase\expandafter{\romannumeral2} with the guidance of \uppercase\expandafter{\romannumeral1}.

\subsection{Loss function}
Following~\cite{hu2019mask}, we combined four losses: identity loss $L_{identity}$~\cite{taigman2016unsupervised}, cycle-consistency loss $L_{cycle}$~\cite{zhu2017unpaired}, adversarial loss $L_{GAN}$~\cite{goodfellow2014generative}, and colour loss for training the proposed network, \textit{i.e.},
\begin{equation}
\begin{aligned}
    \mathcal{L}(G_f,G_s,D^{\mathit{Lab}}_f,&D^{\mathit{Lab}}_s)\\
    =\omega_1 L_{identity}+&\omega_2 L_{cycle}+\omega_3 L_{GAN}+\omega_4 L_{colour}.
    \label{equ:loss}
\end{aligned}
\end{equation}
We follow~\cite{hu2019mask} and empirically set $\omega_1$, $\omega_2$, $\omega_3$, and $\omega_4$ as 5, 10, 1, and 10, respectively.
The generators and discriminators are obtained by solving the mini-max game
\begin{equation}
   \mathit{arg} \min\limits_{G_f,G_s}\max\limits_{D^{\mathit{Lab}}_f,D^{\mathit{Lab}}_s} \mathcal{L}(G_f,G_s,D^{\mathit{Lab}}_f,D^{\mathit{Lab}}_s).
\end{equation}
We define the four losses in Eq.~(\ref{equ:loss}) as follows.

\noindent\textbf{Identity loss} encourages $\overline{I}^{\mathit{Lab}}_s$ and $\overline{I}^{\mathit{Lab}}_f$ to be the same as $I^{\mathit{Lab}}_s$ and $I^{\mathit{Lab}}_f$, respectively: 
\begin{equation}
\begin{aligned}
    L_{identity}&(G_s,G_f)=L^{s}_{identity}(G_s)+L^{f}_{identity}(G_f)\\
    &=\mathbb{E}_{I_s\sim p(I_s)}\big[\left \|G_s(I_s,M_0),I^{\mathit{Lab}}_s\right \|_1\big]\\
    &+\mathbb{E}_{I_f\sim p(I_f)}\big[\left \|G_f(I_f),I^{\mathit{Lab}}_f\right \|_1\big],
\end{aligned}
\end{equation}
where $\left \|.\right\|_1$ represents $L_1$ loss.

\noindent\textbf{Cycle-consistency loss} encourages $\widetilde{I}^{Lab}_{s}$ and $\widetilde{I}^{Lab}_{f}$ to be the same as $I^{\mathit{Lab}}_s$ and $I^{\mathit{Lab}}_f$, respectively:
\begin{equation}
\begin{aligned}
    L_{cycle}&=L^{s}_{cycle}(G_f,G_s)+L^{f}_{cycle}(G_s,G_f)\\
    &=\mathbb{E}_{I_s\sim p(I_s)}\big[\left\|G_s(G_f(I_s),M_g),I^{\mathit{Lab}}_s\right \|_1\big]\\
    &+\mathbb{E}_{I_f\sim p(I_f)}\big[\left\|G_f(G_s(I_f,M_r)),I^{\mathit{Lab}}_f\right \|_1\big].
\end{aligned}
\end{equation}

\noindent\textbf{Adversarial loss} matches the real $Lab$ data distribution and the generated $Lab$ data distribution:
\begin{equation}
\begin{aligned}
    L_{GAN}&(G_s,G_f,D^{\mathit{Lab}}_s,D^{\mathit{Lab}}_f)\\
    &=L^{s}_{GAN}(G_s,D^{\mathit{Lab}}_s)+L^{f}_{GAN}(G_f,D^{\mathit{Lab}}_f)\\
    &=\mathbb{E}_{I^{\mathit{Lab}}_s\sim p(I_s)}[\log(D^{\mathit{Lab}}_s(I^{\mathit{Lab}}_s))]\\
    &+\mathbb{E}_{I_f\sim p(I_f)}[\log(1-D^{\mathit{Lab}}_s(G_s(I_f,M_r))]\\
    &+\mathbb{E}_{I^{\mathit{Lab}}_f\sim p(I_f)}[\log(D^{\mathit{Lab}}_f(I^{\mathit{Lab}}_f))]\\
    &+\mathbb{E}_{I_s\sim p(I_s)}[\log(1-D^{\mathit{Lab}}_f(G_f(I_s))].
\end{aligned}
\end{equation}
\noindent\textbf{Colour loss} encourages the colour in $\widetilde{I}^{Lab}_{s}$ and $\widetilde{I}^{Lab}_{f}$ to be the same as $I^{\mathit{Lab}}_s$ and $I^{\mathit{Lab}}_f$, respectively:
\begin{equation}
\label{angle}
\begin{aligned}
    L_{colour}&=L^{s}_{colour}(G_f,G_s)+L^{f}_{colour}(G_s,G_f)\\
    &=\sum_{p}(J-\cos<(G_s(G_f(I_s),M_g))_p,(I^{\mathit{Lab}}_s)_p>)\\
    &+\sum_{p}(J-\cos<(G_f(G_s(I_f,M_r)))_p,(I^{\mathit{Lab}}_f)_p>),
\end{aligned}
\end{equation}
where $()_p$ represents a pixel, $J$ denotes an all-ones matrix with the same size as the input image, and $\cos<,>$ represents the cosine angle between vectors.
Each pixel of the generated image or input image is regarded as a 3D vector that represents the $Lab$ colour.
The cosine angle between two colour vectors equals to 1 when the vectors have the same direction.
The only difference between our loss and the colour loss formulated in~\cite{wang2019under} is the latter calculates the angle rather than the cosine angle of each pixel, in which the angle between two colour vectors equals to 0 when the vectors have the same direction.
We choose this colour loss because the colour loss in~\cite{wang2019under} leads to gradient explosion in our experiments, resulting in training failure.

\subsection{Network details}
The architectures of Module~\uppercase\expandafter{\romannumeral1} and~\uppercase\expandafter{\romannumeral2} are based on Mask-ShadowGAN~\cite{hu2019mask}, which has two generators and two discriminators.
Each generator contains three convolutional layers for input and down-sampling operations, followed by nine residual blocks with the stride-two convolutions and another three convolutional layers for up-sampling and output operations. 
The residual blocks~\cite{he2016deep} are derived from~\cite{johnson2016perceptual} following the architecture of~\cite{gross2016training}, which has been successfully used for style transfer and super-resolution tasks. 
Discriminators are based on PatchGAN~\cite{isola2017image}.
Instance normalisation~\cite{ulyanov2016instance} is used after each convolution layer.
While the general structure of the backbone network is the same as Mask-ShadowGAN, the number of parameters is different.
The original architecture of Mask-ShadowGAN is drawn from CycleGAN~\cite{zhu2017unpaired}, which 
is designed for general image-to-image translation instead of specifically for shadow removal.
In our backbone network, we follow the principle of SqueezeNet~\cite{iandola2016squeezenet} to reduce the channels of Mask-ShadowGAN by half to consider both performance and efficiency.

\begin{figure}[htbp]
    \centering
    \includegraphics[width=0.6\linewidth]{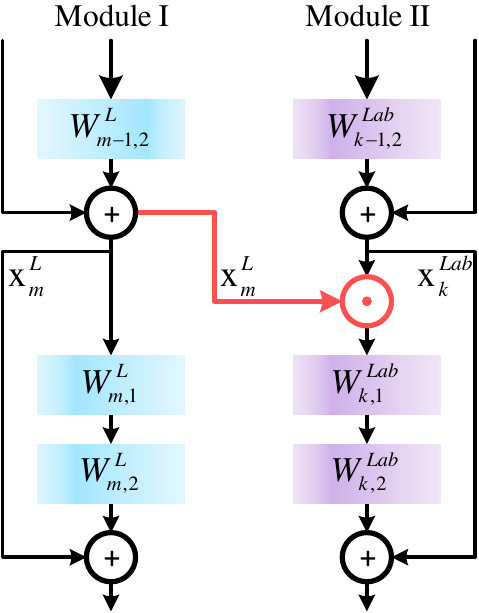}
    \caption{An illustration of the multiplicative connections between the two modules.}
    \label{fig:multi}
\end{figure}
\subsection{Multiplicative connections}
Figure~\ref{fig:multi} shows the details of multiplicative connections used for lightness guidance between the residual blocks of two modules, in which the feature maps from Module~\uppercase\expandafter{\romannumeral1} are connected with the feature maps of Module~\uppercase\expandafter{\romannumeral2} by element-wise multiplication and then sent to the weight layer of next residual block.
The multiplicative connections do not affect the identity mapping $x^{Lab}_k$ and its calculation can be written as:
\begin{equation}
   x^{\mathit{Lab}}_{k+1}=x^{\mathit{Lab}}_{k}+\mathcal{F}(x^{\mathit{Lab}}_{k}\odot x^{L}_{m},W^{\mathit{Lab}}_{k}),
   \label{equmc}
\end{equation}
where $x^{\mathit{Lab}}_{k}$ is the input of the $k$-th layer of Module~\uppercase\expandafter{\romannumeral2}, the function $\mathcal{F}$ represents the residual mapping to be learned, $\odot$ represents the element-wise multiplication, $x^{L}_{m}$ is the input of the $m$-th layer of Module~\uppercase\expandafter{\romannumeral1} and also the input of the $k$-th layer of Module~\uppercase\expandafter{\romannumeral2}, and $W^{\mathit{Lab}}_{k}$ denotes the weights of the $k$-th layer residual unit in Module~\uppercase\expandafter{\romannumeral2}.

\subsection{Convergence analysis}
The training of LG-ShadowNet consists of two steps, and we start from the first step, \textit{i.e.}, the training of Module~\uppercase\expandafter{\romannumeral1}.
In Module~\uppercase\expandafter{\romannumeral1}, if the generator $G^L_f$ and the discriminator $D^L_f$ have enough capacity, the distribution of generated shadow-free data $p(\widetilde{I}^L_f)$ shall converge to the real shadow-free data $p(I^L_f)$ according to the convergence of GANs~\cite{goodfellow2014generative}.

In the second step, $G^L_f$ provides the lightness information to Module~\uppercase\expandafter{\romannumeral2}, which only brings signal changes to the feature maps of Module~\uppercase\expandafter{\romannumeral2} (as shown in Eq.~(\ref{equmc})).
The $G_f$ in LG-ShadowNet is still trained to match the distribution of the real shadow-free data.
Assuming we obtain the optimal generator $G_f$ and discriminator $D_f$ in LG-ShadowNet, 
since $\widetilde{I}^\mathit{Lab}_f\sim G_f(\widetilde{I}^\mathit{Lab}_f|I_f)$, when $I_f$ is applied to $G_f$, we get $\widetilde{I}^\mathit{Lab}_f$ which has the same distribution as the real shadow-free data $I^\mathit{Lab}_f$, and the distribution of the generated shadow-free data $p(\widetilde{I}^\mathit{Lab}_f)$ converges to the distribution of the real shadow-free data $p(I^\mathit{Lab}_f)$.

Likewise, the generator $G^s$ converges to the estimator of real shadow data $p(I^\mathit{Lab}_s)$, and LG-ShadowNet converges to the estimator of the real shadow and shadow-free data.
In practice, if Module~\uppercase\expandafter{\romannumeral1} provides appropriate lightness information in LG-ShadowNet, the model will converge better; on the contrary, if the lightness information provided by Module~\uppercase\expandafter{\romannumeral1} can not guide the learning of shadow removal, the model will converge worse.

\section{Experiments}
\subsection{Datasets and metrics}
In this section, we validate our approach on three widely used shadow removal datasets: 


1) ISTD~\cite{wang2018stacked}, which contains 1,870 image triplets with 1,330 triplets for training and 540 for testing, where a triplet consists of a shadow image, a shadow mask, and a shadow-free image.
ISTD shows good variety in terms of illumination, shape, and scene;

2) AISTD (Adjusted ISTD).
In~\cite{Le2019Shadow}, original shadow-free images in ISTD are transformed to colour-adjusted shadow-free images via a linear regression method~\cite{Le2019Shadow} to mitigate the colour inconsistency between the shadow and shadow-free image pairs.
Since the adjusted shadow-free images have a great influence on the experimental results, we regard this dataset as a new dataset in the following experiments.
AISTD has the same shadow images, shadow masks, and training/testing data splits as ISTD;

3) USR~\cite{hu2019mask}, which contains 2,445 shadow images with 1,956 images for training and 489 for testing.
It also contains 1,770 shadow-free images for training. This is an unpaired dataset that covers a thousand different scenes with great diversity.
There is no corresponding shadow-free image for the shadow images. 

Following recent works~\cite{hu2019direction,qu2017deshadownet,wang2018stacked,hu2019mask,Le2019Shadow}, we conduct quantitative experiments to evaluate the shadow removal performance on paired datasets ISTD and AISTD by computing the root-mean-square error (RMSE) between the ground truth images and generated shadow removal results in $Lab$ colour space, at the original scale $480 \times 640$.

On the unpaired dataset USR, we conduct the user study to evaluate the visual quality of shadow removal results, because this dataset has no ground truth for computing RMSE~\cite{hu2019mask}.
We recruited five participants with average age of 26.
When comparing two methods, we randomly select 30 test images for each participant.
For each image, he/she compares the shadow-removal results from the two methods and then votes for the better one.
We then count the proportion of the 150 votes that are received by each of the two methods as their relative performance: the higher the proportion, the better its shadow-removal quality. 
The following experimental results achieved by our method on different datasets are trained on corresponding datasets respectively.

\subsection{Implementation details}
Our model is initialised following a zero-mean Gaussian distribution with a standard deviation of $0.02$.
The model is trained by using Adam optimiser~\cite{kingma2014adam} and a mini-batch size is set to 1.
Each sample in the training dataset is resized to $448\times448$ and a random crop of $400\times400$ is used for training which prevents the model from learning spatial priors that potentially exist in the dataset~\cite{niklaus2017video}.

For Module~\uppercase\expandafter{\romannumeral1} of our method, we empirically set the training epochs to 200, 200, and 100 for ISTD, AISTD, and USR, respectively.
For Module~\uppercase\expandafter{\romannumeral2} and the variants to be discussed in later experiments, we empirically set the training epochs to 100 on all datasets.
We follow Mask-ShadowGAN to set a basic learning rate $2\times10^{-4}$ for the first half of epochs and reduce the learning rate to zero with a linear decay in the next half of epochs.

\begin{figure}[htbp]
    \centering
    \subfigure[Default: LG-ShadowNet]{
    \includegraphics[width=0.465\linewidth]{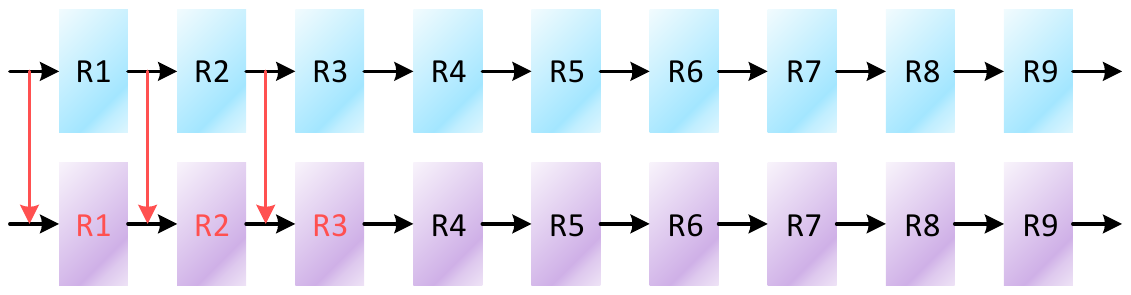}
    \label{fig:mcs}
    }
    \subfigure[Variant: LG-ShadowNet-3]{
    \includegraphics[width=0.465\linewidth]{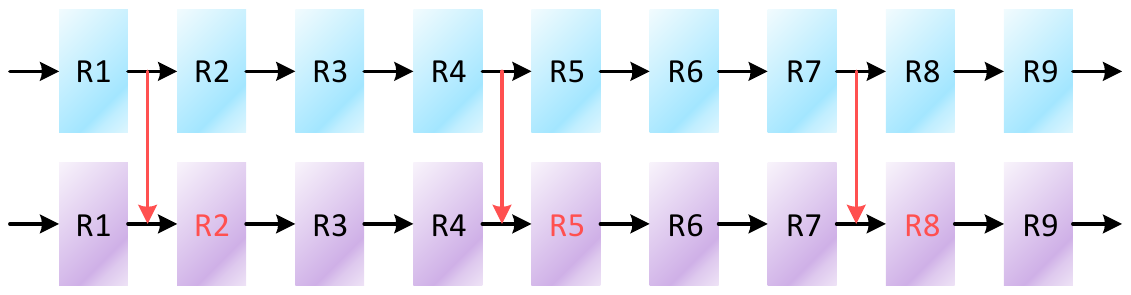}
    \label{fig:mc3}
    }
    \quad
    \subfigure[Variant: LG-ShadowNet-4]{
    \includegraphics[width=0.465\linewidth]{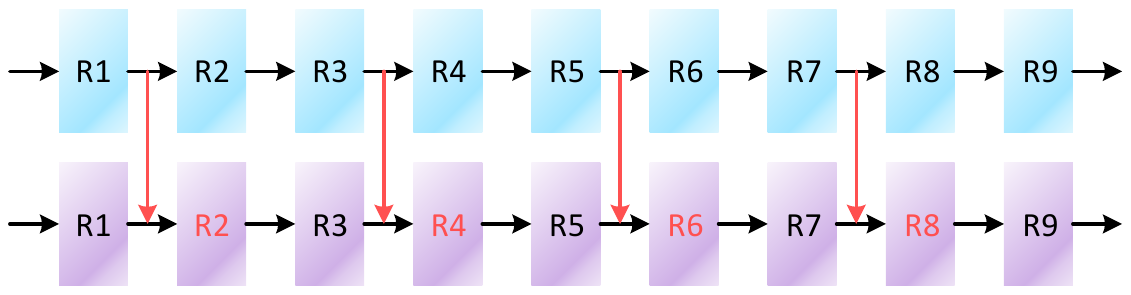}
    \label{fig:mc4}
    } 
    \subfigure[Variant: LG-ShadowNet-9]{
    \includegraphics[width=0.465\linewidth]{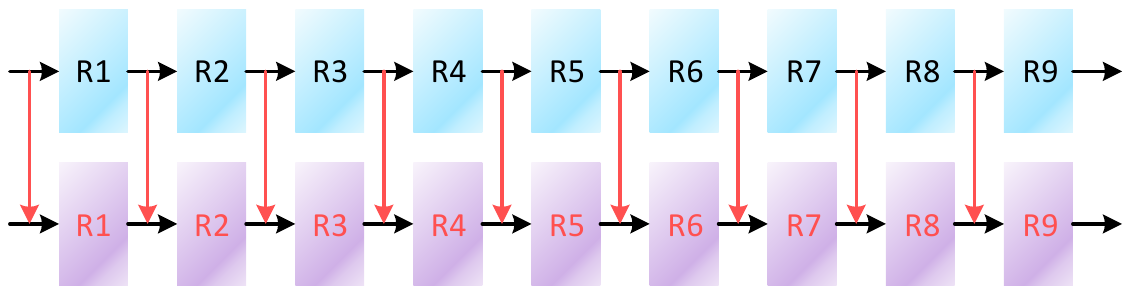}
    \label{fig:mc9}
    }
    \quad
    \subfigure[Variant: LG-ShadowNet-N]{
    \includegraphics[width=0.465\linewidth]{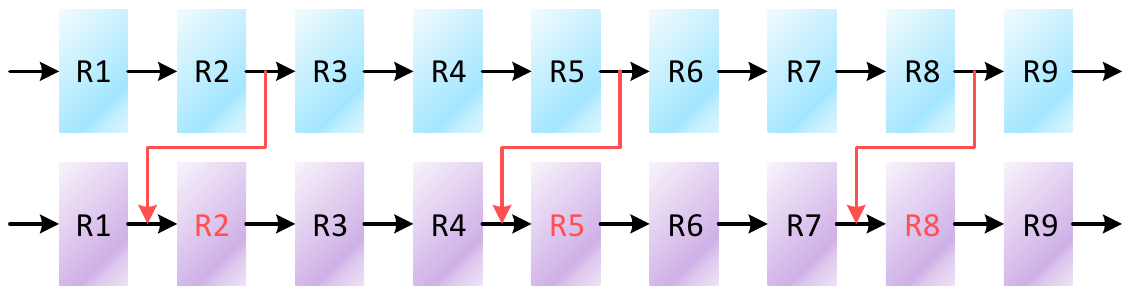}
    \label{fig:mcf}
    } 
    \subfigure[Variant: LG-ShadowNet-P]{
    \includegraphics[width=0.465\linewidth]{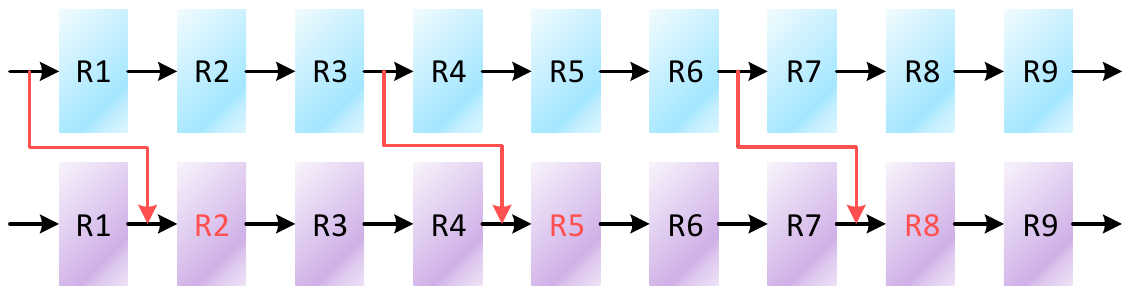}
    \label{fig:mcl}
    }
    \caption{Variants of the multiplicative connections between two modules.
    We show the nine residual blocks (R1-R9) of the generator where the convolutional layers are not shown for simplicity.
    The multiplicative connections are highlighted in red arrows.
    }
    \label{fig:mc}
\end{figure}

Finally, our method is implemented in PyTorch on a computer with a single NVIDIA GeForce GTX 1080 GPU and is evaluated in MATLAB R2016a.
It takes about 68 hours to train LG-ShadowNet on ISTD.

\begin{table*}[htbp]
    \centering
    \caption{Quantitative results on AISTD in terms of RMSE. $Lab$, HSV and RGB indicates the respective colour spaces. S and N represent the RMSE of shadow region and non-shadow region, respectively. `$^{\star}$' denotes training without colour loss, which we use in all the remaining experiments. $L$ and V represent the $L$ channel of $Lab$ and the V channel of HSV, respectively. Module~\uppercase\expandafter{\romannumeral1} is trained to compensate for the lightness on $L$ channel so we only show the RMSE of $L$ channel for evaluating this module.}
    {\begin{tabular}{l|c|ccc|ccc}
    \hline
    \multirow{2}{*}{Method} & \multirow{2}{*}{Training data} &  \multicolumn{6}{c}{AISTD} \\ 
    \cline{3-8} 
    &  & $Lab$ & N & S & $L$ & N & S \\ 
    \hline
    Original data & - & 9.16 & 3.33 & 38.53 & 6.05 & 1.61 & 28.09 \\ 
    \hline
    \multirow{2}{*}{Module~\uppercase\expandafter{\romannumeral1}} & V & - & - & - & 3.70 & 2.26 & 10.73\\
    & $L$ & - & - & - & 3.15 & 2.37 & 7.06\\
    \hline
    \multirow{3}{*}{Module~\uppercase\expandafter{\romannumeral2}$^{\star}$} & RGB & 5.78 & 4.74 & 12.03 & 3.23 & 2.49 & 7.48\\
    & HSV & 7.71 & 6.37 & 15.84 & 3.57 & 2.77 & 7.96\\
    & $Lab$  & 5.64 & 4.66 & 11.65 & 3.35 & 2.71 & 7.24 \\
    \hline
    \multirow{3}{*}{LG-ShadowNet$^{\star}$} & $Lab+Lab$ & 5.40 & 4.33 & 12.04 & 3.11 & 2.39 & 7.42\\
    & V$+Lab$  & 5.29 & 4.25 & 11.27 & 3.04 & 2.31 & 7.05 \\
    & $L+Lab$ & \textbf{5.17} & \textbf{4.09} & \textbf{11.15}  & \textbf{2.98} & \textbf{2.23} & \textbf{6.93} \\
    \hline
    Module~\uppercase\expandafter{\romannumeral1} \textit{Sup.} & $L$ & - & - & - & 2.98 & 2.28 & 6.71\\
    Module~\uppercase\expandafter{\romannumeral2}$^{\star}$ \textit{Sup.} & $Lab$  & 5.08 & 4.07 & 11.28 & 2.88 & 2.22 & 6.83 \\
    LG-ShadowNet$^{\star}$ \textit{Sup.} & $L+Lab$ & \textbf{4.83} & \textbf{3.95} & \textbf{10.07}  & \textbf{2.66} & \textbf{2.10} & \textbf{5.83} \\
    \hline
    \end{tabular}}
    \label{tab:ablation-lightness}
\end{table*}
\begin{figure*}[htbp]
    \centering
    \includegraphics[width=0.8\linewidth]{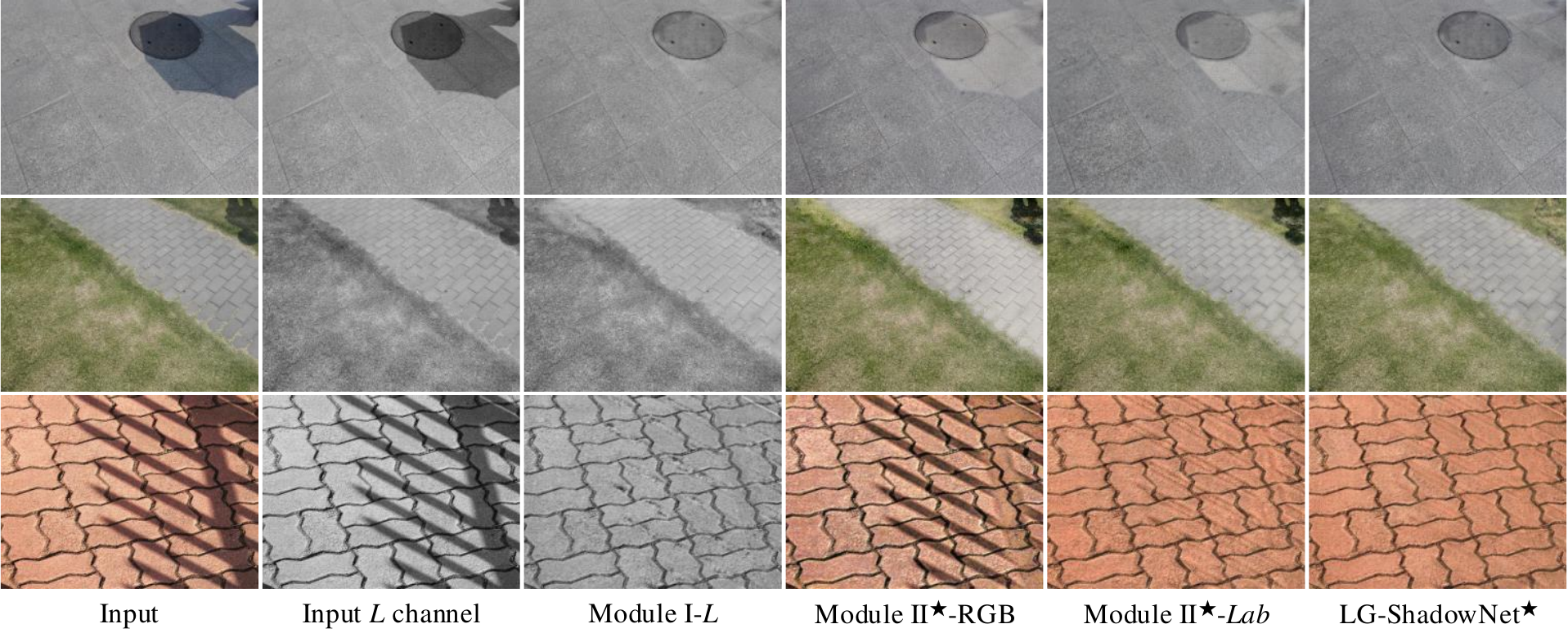}
    \caption{Visual comparisons on ISTD, AISTD and USR.
    Three rows from top to bottom show results for one sample from ISTD, AISTD and USR, respectively.
    The first two columns show the input shadow images and their $L$ channels, respectively.}
    \label{fig:ablation}
\end{figure*}

\subsection{Variants of two-module connections}\label{sec:variants}
To study the impact of using multiplicative connections, we try several variants of connections.
These variants are shown in Fig.~\ref{fig:mc}.
The default connections in LG-ShadowNet, as shown in Fig.~\ref{fig:mcs}, are inserted at the first three shallow residual blocks.
No connections are inserted in middle or deeper blocks since they have been shown to hurt the performance~\cite{yosinski2014transferable}.
The following three variants shown in Fig.~\ref{fig:mc3}-\ref{fig:mc9} connect corresponding layers in both modules, which indicates the case that $k = m$ in Eq.~(\ref{equmc}).
These three variants show different intervals and different numbers of connections.
Specifically, LG-ShadowNet-3 inserts connections after the first of every three residual blocks, \textit{i.e.}, with an interval of three, similar to the setting in the Spatiotemporal Multiplier Networks~\cite{feichtenhofer2017spatiotemporal}.
Similarly, LG-ShadowNet-4 and LG-ShadowNet-9 insert the connections between the residual blocks with an interval of two and one, respectively.
The two variants of LG-ShadowNet-N and LG-ShadowNet-P, shown in Fig.~\ref{fig:mcf}-\ref{fig:mcl},
connect between non-corresponding residual layers, \textit{i.e.}, $k\neq m$ in Eq.~(\ref{equmc}).
They insert the connections from the next ($(k+1)$-th) and previous ($(k-1)$-th) residual blocks of Module~\uppercase\expandafter{\romannumeral1} to the current ($k$-th) residual blocks of Module~\uppercase\expandafter{\romannumeral2}, respectively.

We also study the impact of additive connections, \textit{i.e.}, using addition instead of the element-wise multiplication with $k=m$ in Eq.~(\ref{equmc}) in LG-ShadowNet.
We denote such variant network as LG-ShadowNet-A. 

\subsection{Ablation study}
We first perform an ablation study on AISTD to evaluate the effectiveness of the proposed lightness-guided architecture trained with unpaired data in different colour spaces.
We use the default connection in LG-ShadowNet and remove the proposed colour loss to train different models with different inputs with different number of input channels.
Note that Module~\uppercase\expandafter{\romannumeral2} is trained by following the settings of LG-ShadowNet without the guidance of Module~\uppercase\expandafter{\romannumeral1} and the colour loss. 
Besides, to verify the effectiveness of the lightness-guided architecture trained on paired data, we use a generator that maps shadow-data to shadow-free data and the $L_1$ loss to train Modules~\uppercase\expandafter{\romannumeral1} and~\uppercase\expandafter{\romannumeral2} on paired data in a \textit{fully supervised} manner, and these models are denoted with a suffix \textit{Sup.}.
Quantitative results in terms of RMSE metric are shown in Table~\ref{tab:ablation-lightness} and the RMSE between the real shadow and shadow-free pairs on AISTD testing sets
is shown in the first row of the table.

\begin{table*}[htbp]
    \centering
    \caption{Quantitative results on ISTD, AISTD and USR.
    Each result on USR represents the proportion of votes received by the proposed LG-ShadowNet$^{\star}$ or its variants when compared with Module~\uppercase\expandafter{\romannumeral2}$^{\star}$ trained on $Lab$ or RGB data.
    The suffix `-$Lab$' and `-RGB' in the model name represent the training data on $Lab$ and RGB colour spaces, respectively, which we use in all the remaining experiments.}
    {\begin{tabular}{l|ccc|ccc|cc}
    \hline
    \multirow{2}{*}{Method} & \multicolumn{3}{c|}{ISTD} & \multicolumn{3}{c|}{AISTD} & \multicolumn{2}{c}{USR} \\ 
    \cline{2-9} 
     & $Lab$ & N & S & $Lab$ & N & S & Module~\uppercase\expandafter{\romannumeral2}$^{\star}$-$Lab$ & Module~\uppercase\expandafter{\romannumeral2}$^{\star}$-RGB \\ 
    \hline
    LG-ShadowNet$^{\star}$ & 6.64 & 6.00 & 10.98 & \textbf{5.17} & \textbf{4.09} & 11.15 & 64.7\% & 88.7\%\\
    \hline
    LG-ShadowNet-3$^{\star}$ & 6.80 & 6.19 & \textbf{10.70}  & 5.34 & 4.26 & \textbf{11.11} & 61.3\% & 91.3\%\\
    LG-ShadowNet-4$^{\star}$ & 6.69 & 6.06 & 10.78 & 5.37 & 4.26 & 11.17 & 61.3\% & 87.3\% \\
    LG-ShadowNet-9$^{\star}$ & 6.91 & 6.26 & 11.34 & 5.46 & 4.28 & 11.34 & \textbf{72.0\%} & \textbf{92.7\%} \\
    \hline
    LG-ShadowNet-N$^{\star}$ & \textbf{6.61} & \textbf{5.94} & 10.93 & 5.39 & 4.29 & 11.18 & 64.0\% & 90.7\% \\
    LG-ShadowNet-P$^{\star}$ & 6.80 & 6.14 & 11.30  & 5.39 & 4.28 & 11.37 & 54.7\%  & 88.7\% \\
    \hline
    LG-ShadowNet-A$^{\star}$ & 6.89 & 6.13 & 11.58 & 5.26 & \textbf{4.09} & 11.57 & 52.0\% & 79.3\% \\
    \hline
    \end{tabular}}
    \label{tab:ablation}
\end{table*}
\begin{table}[htbp]
    \centering
    \caption{Quantitative results of LG-ShadowNet trained with and without the colour loss on ISTD, AISTD and USR.}
    {\begin{tabular}{c|ccc|ccc|c}
    \hline
    \multirow{2}{*}{Method}  & \multicolumn{3}{c|}{ISTD} & \multicolumn{3}{c|}{AISTD} & USR\\ 
    \cline{2-8} 
    & $Lab$ & N & S & $Lab$ & N & S & $Lab$ \\ 
    \hline
    w/o $L_{colour}$ & \textbf{6.64} & 6.00 & \textbf{10.98} & 5.17 & 4.09 & 11.15 & 36.7\%\\
    \hline
    with $L_{colour}$ & 6.67 & \textbf{5.91} & 11.63 & \textbf{5.02} & \textbf{4.02} & \textbf{10.64} & \textbf{63.3}\%\\
    \hline
    \end{tabular}}
    \label{tab:colour}
\end{table}

From the rows 2 and 3 of Table~\ref{tab:ablation-lightness}, we can see that Module~\uppercase\expandafter{\romannumeral1} trained on
V channel and $L$ channel can significantly reduce the RMSE of the original data (row 1).
The latter demonstrates the effectiveness of using $L$ channel for lightness compensation.
The results in rows 4-6 show that, using data on $Lab$ colour space as training data is more suitable for shadow removal than using RGB or HSV colour spaces.
The results in rows 7-9 show that, using $L$ channel as the input of Module~\uppercase\expandafter{\romannumeral1} achieves the best results than using the V channel and the $Lab$ data.
This confirms that the benefits are from the learned lightness information and the superiority of using $L$-channel data to guide the learning of shadow removal.
Compared with Module~\uppercase\expandafter{\romannumeral2}$^{\star}$ trained on $Lab$ data (row 6), 
LG-ShadowNet$^{\star}$ trained on $L+Lab$ data (row 9) can reduce RMSE by 8.3\% from 5.64 to 5.17, which proves the effectiveness of the lightness-guided architecture.
In addition, the results in rows 10-12 show that training above modules on paired data in a fully supervised manner can further improve the performance.
The advantage of using $L$ channel as the guidance and the effectiveness of the lightness-guided architecture are further verified here. 

Figure~\ref{fig:ablation} shows some visual comparison results of Module~\uppercase\expandafter{\romannumeral1} trained on $L$ data (Module~\uppercase\expandafter{\romannumeral1}-$L$), Module~\uppercase\expandafter{\romannumeral2}, and LG-ShadowNet$^{\star}$ on the ISTD, AISTD and USR datasets.
From the second and third columns, we can see that Module~\uppercase\expandafter{\romannumeral1}-$L$ can restore the shadow regions on $L$ channel effectively.
LG-ShadowNet$^{\star}$ can produce better results than individual modules, \textit{e.g.}, it successfully removes the shadow and restores the lightness on the top right of the image shown in the second row.

Next, we perform another ablation study on ISTD, AISTD and USR to evaluate the various connection variants of LG-ShadowNet described in subsection~\ref{sec:variants}.
All models are trained on unpaired $L+Lab$ data without the proposed colour loss.
Quantitative results in terms of RMSE metric on ISTD and AISTD and the user study results of LG-ShadowNet and its variants against Module~\uppercase\expandafter{\romannumeral2} on USR are shown in Table~\ref{tab:ablation}.

From Tables~\ref{tab:ablation}, we can see that different variants achieve different results on different datasets.
We observe that LG-ShadowNet-N$^{\star}$ achieves better performance on the ISTD dataset. This indicates that inserting multiplicative connections between deeper layers of Module~\uppercase\expandafter{\romannumeral1} to shallower layers of Module~\uppercase\expandafter{\romannumeral2} could be more effective.

On AISTD, LG-ShadowNet$^{\star}$ significantly surpasses other variants, which shows that low-level features are sufficient for guiding the learning of shadow removal, while embedding more high-level features leads to inferior results, especially in non-shadow regions.
The variant of LG-ShadowNet-9$^{\star}$ achieves the best result on USR, which means using the lightness features from deeper layers is more effective in producing visually pleasing results.
Compared LG-ShadowNet$^{\star}$ with the LG-ShadowNet-A$^{\star}$, we observe that the additive connections lead to inferior performance.
This may be due to the fact that multiplicative interactions can bring stronger signal change than additive interactions~\cite{feichtenhofer2017spatiotemporal}.
We chose LG-ShadowNet as the default connection variant because it performs more robust than other variants on all three datasets.

Finally, we report the qualitative results of LG-ShadowNet trained with and without the proposed colour loss in Table~\ref{tab:colour} to evaluate the effectiveness of the proposed colour loss.
On the ISTD dataset, we can see that the colour loss has little effect on the overall RMSE, but it improves the RMSE of non-shadow regions, \textit{i.e.}, the quality of most parts of the results is improved.
Comparing the statistics on AISTD and USR, we observe the conspicuous improvement by using the proposed colour loss.
On the whole, the colour loss that restricts the colour direction to be the same is an effective constraint for shadow removal.

\subsection{Comparison with the state-of-the-arts}

\begin{figure}[htbp]
    \centering
    \includegraphics[width=1\linewidth]{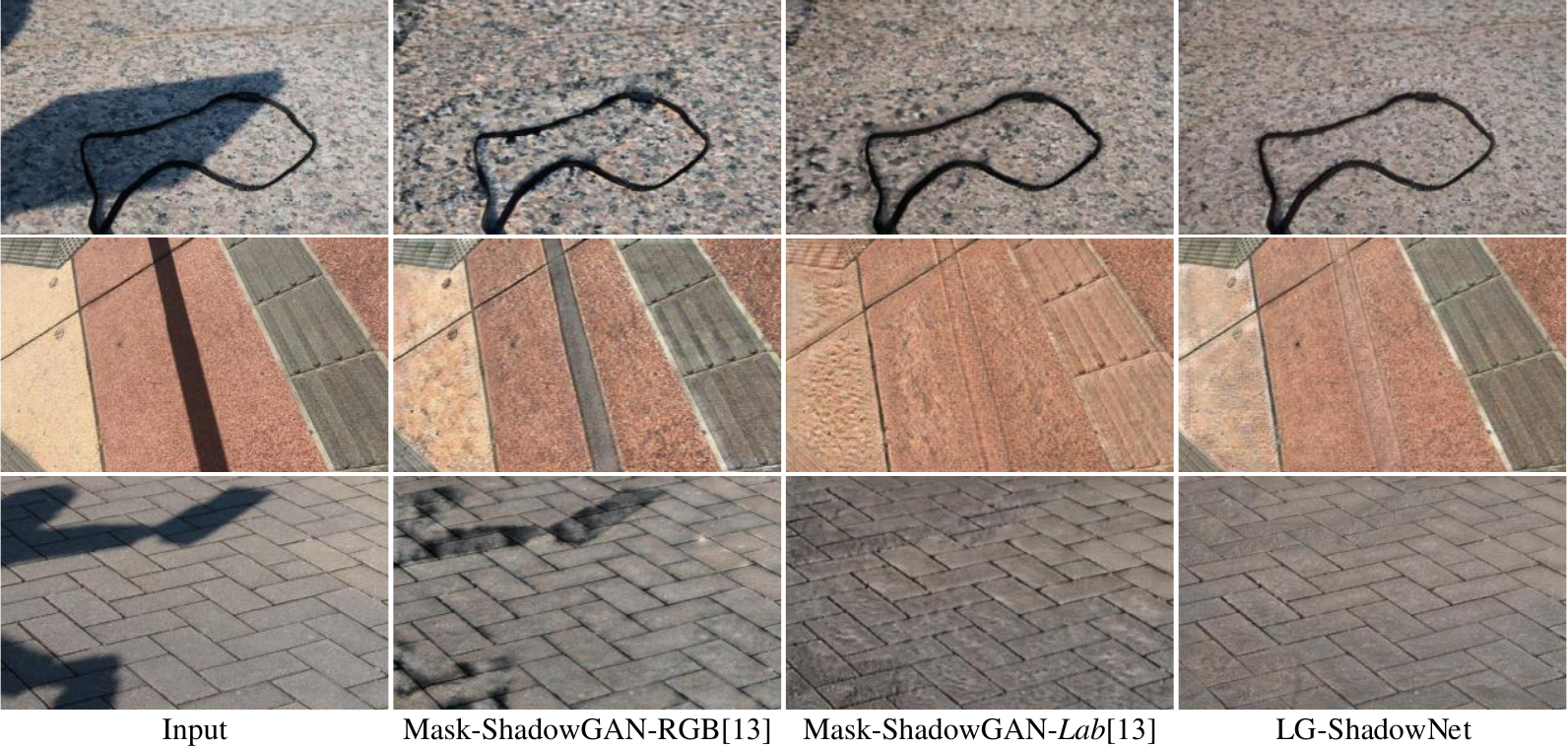}
    \caption{Visual comparisons on USR. Each row shows results for one sample image.}
    \label{fig:usr}
\end{figure}
\begin{table*}[htbp]
    \centering
    \caption{Quantitative results on ISTD and AISTD in terms of RMSE. `-' denotes the result is not publicly reported.
    ARGAN~\cite{ding2019argan} uses their own way to calculate RMSE, which is different from other methods, and their results are denoted with `$^*$'.}
    {\begin{tabular}{c|c|c|ccc|ccc}
    \hline
    \multirow{2}{*}{Data type} & \multirow{2}{*}{Method} & \multirow{2}{*}{Training}  & \multicolumn{3}{c|}{ISTD} & \multicolumn{3}{c}{AISTD}\\ 
    \cline{4-9} 
    & & & $Lab$ & N & S & $Lab$ & N & S \\ 
    \hline
    \multirow{3}{*}{Prior-based}
                                & Yang \textit{et al.}~\cite{yang2012shadow} & N/A & 15.63 & 14.83 & 19.82 & 16.80 & 15.20 & 25.30 \\
                                & Guo \textit{et al.}~\cite{guo2012paired} & N/A & 9.30 & 7.46 & 18.95 & 7.10 & 4.30 & 22.30 \\
                                & Gong \textit{et al.}~\cite{gong2014interactive} & N/A & 8.53 & 7.29 & 14.98 & 5.10 & 3.40 & 14.40 \\
    \hline
    \multirow{4}{*}{Paired}     
                                & ST-CGAN~\cite{wang2018stacked} & RGB & 7.47 & 6.93 & 10.33 & 9.50 & 8.60 & 14.00 \\
                                & DSC~\cite{hu2019direction} & $Lab$ & 6.67 & 6.14 & 9.76 & - & - & - \\
                                & ARGAN~\cite{ding2019argan} & RGB & 6.68$^*$ & 5.83$^*$ & 7.21$^*$ & - & - & - \\
                                & SP+M-Net~\cite{Le2019Shadow} & RGB & - & - & - & 4.41 & 3.64 & 8.84
                                \\
    \hline
    \multirow{4}{*}{Unpaired}   & CycleGAN~\cite{zhu2017unpaired} & RGB & 8.16 & - & - & - & - & - \\
                                & Mask-ShadowGAN~\cite{hu2019mask} & RGB & 7.41 & 6.68 & 12.67 & 5.48 & 4.52 & 11.53 \\
                                & Mask-ShadowGAN~\cite{hu2019mask} & $Lab$ & 7.32 & 6.57 & 12.65 & 5.84 & 4.82 & 12.28\\
                                & LG-ShadowNet & $L+Lab$ & \textbf{6.67} & \textbf{5.91} & \textbf{11.63} & \textbf{5.02} & \textbf{4.02} & \textbf{10.64} \\
    \hline
    \end{tabular}}
    \label{tab:all}
\end{table*}

In this subsection, we compare our full model with several state-of-the-art methods on the ISTD, AISTD and USR datasets. 
Results of Mask-ShadowGAN are obtained by training and testing on each dataset using the code provided by its authors, while other results are provided by the authors of ST-CGAN~\cite{wang2018stacked}, DSC~\cite{hu2019direction} and SP+M-Net~\cite{Le2019Shadow}.

First of all, we compare our method with Mask-ShadowGAN on the USR dataset through the user study, where Mask-ShadowGAN trained on RGB data reports the most recent state-of-the-art performance.
The proportions of votes received by LG-ShadowNet when compared with Mask-ShadowGAN trained on RGB and $Lab$ data are 80.7\% and 72.7\%, respectively.
These results show that, after converting the input data from RGB to $Lab$, Mask-ShadowGAN actually performs even better on USR.
However, the proposed LG-ShadowNet still receives more votes than Mask-ShadowGAN trained on RGB or $Lab$ data.
Qualitative results are shown in Fig.~\ref{fig:usr}.

\begin{figure*}[htbp]
    \centering
    \includegraphics[width=1\linewidth]{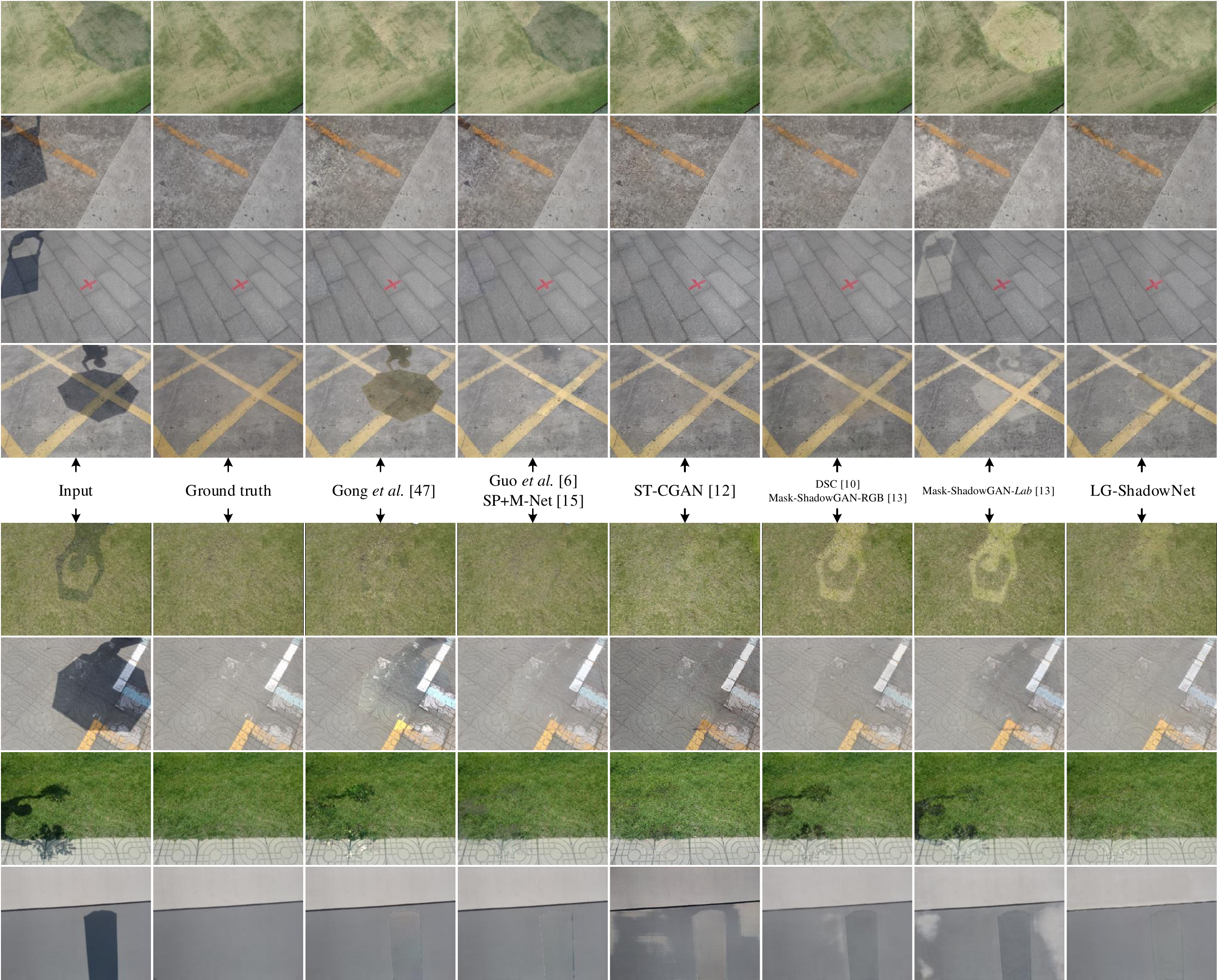}
    \caption{Visual comparisons on ISTD and AISTD. Each row shows results for one sample image.}
    \label{fig:istd}
\end{figure*}

Next, we compare the proposed method with the state-of-the-art methods on the ISTD and AISTD datasets.
Among them, Gong \textit{et al.}~\cite{gong2014interactive}, Guo \textit{et al.}~\cite{guo2012paired}, and Yang \textit{et al.}~\cite{yang2012shadow} remove shadows based on image priors. ST-CGAN~\cite{wang2018stacked}, DSC~\cite{hu2019direction}, and ARGAN~\cite{ding2019argan} are trained using paired shadow and shadow-free images.
SP+M-Net~\cite{Le2019Shadow} is trained by using shadow and shadow-free image pairs, as well as shadow masks.
CycleGAN~\cite{zhu2017unpaired} and Mask-ShadowGAN~\cite{hu2019mask} are trained using unpaired images.

The quantitative results are shown in Table~\ref{tab:all}.
we can see that our method outperforms the methods based on image priors and those using unpaired data on both datasets.
Compared with the methods using paired data, our method is also competitive, achieving comparable results to DSC~\cite{hu2019direction} on the ISTD dataset.
Note that Module~\uppercase\expandafter{\romannumeral2}$^{\star}$ trained on $Lab$ data performs better than Mask-ShadowGAN on AISTD, and the former has fewer parameters.
This proves the effectiveness of using the strategy of SqueezeNet~\cite{iandola2016squeezenet} to reduce the model parameters.

Figure~\ref{fig:istd} shows the qualitative results of LG-ShadowNet and several state-of-the-art methods on four challenging sample images in the ISTD (rows 1-4) and AISTD (rows 5-8) datasets.
Compared with Mask-ShadowGAN, LG-ShadowNet better restores the lightness of the shadow regions on all samples.
Our method is also comparable to the methods using paired data, especially on the samples in ISTD. 
It is worth noting that our method can better deal with the shadow edges than SP+M-Net (rows 5-8 and column 4).
The reason is that our method uses the continuous lightness information to guide the shadow removal while SP+M-Net uses binary shadow masks.
These visual results verify the effectiveness of the proposed method for shadow removal. 


\section{Conclusion}
In this paper, we proposed a new lightness-guided method for shadow removal using unpaired data.
It fully explores the important lightness information by first training a CNN module only for lightness before considering other colour information.
Another CNN module is then trained with the guidance of lightness information from the first CNN module to integrate the lightness and colour information for shadow removal.
A colour loss is proposed to further utilise the colour prior of existing data.
Experimental results demonstrate that the effectiveness of the proposed lightness-guided architecture and our LG-ShadowNet outperforms the state-of-the-art methods with training on unpaired data.

\section*{Acknowledgement}

This work was supported by National Nature Science Foundation of China (51827813, 61472029, 61672376, U1803264, 61473031), R\&D Program of Beijing Municipal Education commission and the Fundamental Research Funds for the Central Universities (2018YJS045).
The authors would like to thank Mr.~Yang~Yu and his team for their remarkable works on user study.

\ifCLASSOPTIONcaptionsoff
  \newpage
\fi



\bibliographystyle{IEEEtran}
\bibliography{Shadowremoval}

\end{document}